\documentclass[
]{ceurart}

\usepackage{listings}
\usepackage{comment}
\usepackage{algorithm}
\usepackage{algpseudocode}
\usepackage{mathtools}
\usepackage{comment}
\usepackage{enumitem}
\usepackage{multirow}
\usepackage{colortbl}

\newtheorem{theorem}{Theorem}[section]
\newtheorem{proposition}[theorem]{Proposition}

\newtheorem{definition}{Definition}[section]

\newcommand{\WRP}{\par\qquad\(\hookrightarrow\)\enspace}

\begin{document}


\copyrightyear{2023}
\copyrightclause{Copyright for this paper by its authors.
  Use permitted under Creative Commons License Attribution 4.0
  International (CC BY 4.0).}

\conference{MUWS'23: 2nd International Workshop on Multimodal Human Understanding for the Web and Social Media, October 22, 2023, Birmingham, UK}

\title{Towards Subject Agnostic Affective Emotion Recognition}

%\tnotetext[1]{You can use this document as the template for preparing your
%  publication. We recommend using the latest version of the ceurart style.}


\author[1]{Amit Kumar Jaiswal}[%
orcid=0000-0001-8848-7041,
email=a.jaiswal@surrey.ac.uk,
%url=https://yamadharma.github.io/,
]
%\cormark[1]
\address[1]{University of Surrey, United Kingdom}

\author[2]{Haiming Liu}[%
orcid=0000-0002-0390-3657,
email=h.liu@soton.ac.uk,
%url=https://kmitd.github.io/ilaria/,
]
\address[2]{University of Southampton, United Kingdom}

%3rd author
\author[3]{Prayag Tiwari}[%
orcid=0000-0002-2851-4260,
email=prayag.tiwari@ieee.org,
%url=https://kmitd.github.io/ilaria/,
]
\address[3]{School of Information Technology, Halmstad University, Sweden}

%\cortext[1]{Corresponding author.}
%\fntext[1]{These authors contributed equally.}

\begin{abstract}
This paper focuses on affective emotion recognition, aiming to perform in the subject-agnostic paradigm based on EEG signals. However, EEG signals manifest subject instability in subject-agnostic affective Brain-computer interfaces (aBCIs), which led to the problem of distributional shift. Furthermore, this problem is alleviated by approaches such as domain generalisation and domain adaptation. Typically, methods based on domain adaptation confer comparatively better results than the domain generalisation methods but demand more computational resources given new subjects. We propose a novel framework, meta-learning based augmented domain adaptation for subject-agnostic aBCIs. Our domain adaptation approach is augmented through meta-learning, which consists of a recurrent neural network, a classifier, and a distributional shift controller based on a sum-decomposable function. Also, we present that a neural network explicating a sum-decomposable function can effectively estimate the divergence between varied domains. The network setting for augmented domain adaptation follows meta-learning and adversarial learning, where the controller promptly adapts to new domains employing the target data via a few self-adaptation steps in the test phase. Our proposed approach is shown to be effective in experiments on a public aBICs dataset and achieves similar performance to state-of-the-art domain adaptation methods while avoiding the use of additional computational resources.
\end{abstract}


\maketitle

\section{Introduction}
The human emotional experience and the understanding of its intricate interplay can be a challenging task. Recent technological advancements in brain-computer interaction systems, specifically affective Brain-computer interfaces, have enabled the automatic identification of user emotions and facilitated a more humanised mode of interaction~\cite{muhl2014survey,shanechi2019brain}. The electroencephalography (EEG) signal, in particular, has been utilised in subject-dependent emotion models to recognise user emotions, with both the training and test data derived from a single subject~\cite{jenke2014feature,wang2022systematic}. Despite the potential of EEG signals in aBCIs, their application is limited by the non-stationary nature of these signals and the structural variability among different subjects impose significant challenges in the development of subject-independent models. These models are often following the assumption of independent and identically distributed samples\footnote{https://en.wikipedia.org/wiki/Independent\_and\_identically\_distributed\_random\_variables} and typically exhibit poor generalisation performance in practical aBCI applications due to the problem of domain shift~\cite{sugiyama2007covariate,samek2013transferring,sussillo2016making,lin2017improving,fdez2021cross}. 
To address the problem of domain shift in subject-agnostic EEG-based emotion recognition, an emergent approach of Domain adaptation can be leveraged, which utilises data from both the source and target domains to enhance the adaptation performance. A key domain adaptation technique involves mapping the two distributions to a shared feature space, where they have identical marginal distributions. Despite its considerable success in subject-independent EEG-based emotion recognition~\cite{lin2017improving,lan2018domain,fdez2021cross}, domain adaptation approaches can be computationally intensive and time-consuming, which poses a vexing challenge leading to suboptimal user experiences in real-world applications. To address this issue, the notion of domain generalisation has emerged, particularly in scenarios in which multiple source domains are accessible with a lack of unlabelled target samples. The subject-agnostic emotion recognition models can be constructed using domain generalisation techniques~\cite{lew2020eeg,xu2022dagam}. Nonetheless, given that there is no prior knowledge pertaining to the target domain during training, it becomes arduous for domain generalisation to achieve performance on par with that domain adaptation. A potential approach is to leverage adaptive subspace feature matching (ASFM), which pre-trains the primary model and utilises a limited number of test samples to adjust efficiently~\cite{chai2017fast}. While the ASFM approach can evade the time-consuming nature of adaptation, most of them necessitate the retention of both source and target domains in the test phase, which leads to additional storage requirements and reduces portability~\cite{gu2022multi}. Nevertheless, in real-world applications, the ability of an EEG-based affective model to rapidly adapt to different subjects while maintaining its portability is crucial. 

This paper presents a novel framework, namely, meta-learning based augmented domain adaptation (MeLaDA) for subject-agnostic EEG-based emotion recognition. Unlike the traditional adaptive subspace feature matching (ASFM), MeLaDA only demands the target domain during the test phase. Therefore, MeLaDA can generate predictions more rapidly compared to domain adaptation and ASFM. Based on the viewpoint of real-world applications, MeLaDA is better suited to constructing emotion models for subject-agnostic aBCIs. The proposed meta-learning based augmented domain adaptation (MeLaDA) framework is implemented by formulating the equivalence of a network with a sum-decomposable structure to domain discrepancy metrics utilised in classical domain adaptation techniques such as maximum mean discrepancy~\cite{pan2011domain,long2017deep} or $\mathcal{H}$-divergence~\cite{ben2010theory,zhao2019learning}. Using our formulation, we present the MeLaDA framework, which incorporates a classifier, a feature extractor, and a sum-decomposable structure termed domain shift regulator. By leveraging the benefits of adversarial learning and meta-learning, the regulator facilitates the MeLaDA model's rapid generalisation to new domains by using the target data through a few self-adaptive steps during the test phase. The key contributions of our approach are three-fold:
\begin{enumerate}[leftmargin=*]
    \item MeLaDA derives a pertinent approach to develop a subject-agnostic EEG-based emotion recognition model and a way to incorporate any type of domain discrepancy explicated by a sum-decomposable network.
    \item Our proposed framework is constructed to be portable and able to quickly adapt to various subjects for EEG-driven emotion recognition.
    \item We carried out extensive experiments on the publicly available EEG-based aBCIs dataset, SEED\footnote{https://bcmi.sjtu.edu.cn/home/seed/seed.html}. The results of the experiments indicate that our proposed approach outperforms domain generalisation methods. Moreover, our proposed approach, MeLaDA, exhibits comparable time and storage costs to domain generalisation methods.
\end{enumerate}

\section{Motivation}
The key motivation behind our MeLaDA framework is to simplify the estimation of domain shift by leveraging a basic network that only requires the target domain as input. This is in contrast to traditional domain adaptation approaches which compare the target domain with a specific source domain, requiring additional storage space for source data and complex methods such as generative adversarial network (GAN)~\cite{goodfellow2020generative} to represent domain shift during the test phase. These limitations make the practical application of domain adaptation approaches difficult in EEG-driven emotion recognition. However, in a multi-source scenario, we demonstrate that minimising the discrepancy between all pairwise domains is equivalent to minimising the discrepancy between each domain and an implicit domain. Moreover, we prove that any domain shift metrics can be represented theoretically by a network with a sum-decomposition form.

\section{Prior Work}
\textbf{EEG-based Emotion Recognition:} The inherent non-stationarity of EEG signals and variability across individuals, developing a subject-independent model for EEG-based emotion recognition using conventional machine learning methods is challenging. Recently, attention has been directed towards affective brain-computer interfaces~\cite{muhl2014survey}, which explicated the concept of aBCIs by integrating affective factors into traditional brain-computer interfaces~\cite{zander2011context}. Subsequently, there has been a focus on the application of aBCIs in EEG-based emotion recognition~\cite{zheng2015investigating} which involves 15 participants who watched selected Chinese movie clips to elicit three emotions i.e., happy, neutral, and sad. They curated an EEG emotion recognition dataset called SEED, in which they recorded the EEG signals of the participants. Building upon the SEED dataset, researchers have made significant advancements in developing models for EEG-based emotion recognition, particularly in the context of subject-dependent models. To address this issue, researchers have turned their attention to domain adaptation and domain generalisation techniques for subject-independent EEG-based emotion recognition. Domain adaptation approaches primarily focus on reducing domain shift by minimising discrepancies between different domains using established metrics such as maximum mean discrepancy (MMD)~\cite{pan2011domain,long2017deep,wang2018visual}, the Kullback-Leibler divergence~\cite{zhuang2015supervised}, and $\mathcal{H}$-divergence~\cite{ben2010theory}. Existing work~\cite{zheng2015investigating} among varied domains that employed transfer component analysis~\cite{pan2011domain} to minimise MMD~\cite{gretton2006kernel} by constructing a kernel matrix, which led to the successful development of personalised EEG-based emotion models. 
\\
\textbf{Domain Adaptation and Domain Generalisation:} Adversarial domain adaptation methods have gained significant attention and emerged as successful approaches across various application~\cite{ganin2015unsupervised,tzeng2017adversarial,shen2018wasserstein}. These methods draw inspiration from the concept of GAN, which involves adversarial training to align the generated distribution with the real distribution. In the field of aBCIs, researchers have also embraced adversarial domain adaptation approaches with successful outcomes. For instance, the usage of domain-adversarial neural networks (DANN)~\cite{ganin2016domain} for EEG-based emotion recognition, in subject-independent models~\cite{li2018cross}. Furthermore, the adoption of Wasserstein GAN~\cite{arjovsky2017wasserstein} for domain adaptation (WGAN-DA) has been successfully utilised for facilitating subject-independent emotion recognition models~\cite{wganda}. From the viewpoint of practical scenarios pertaining to aBCIs, each subject represents an individual domain. Domain adaptation (DA) approaches, although computationally intensive for new domains, have been commonly employed in aBCIs. However, domain generalisation techniques, which generalise to unseen target domains without requiring additional target domain data, have gained traction in aBCIs~\cite{wang2022generalizing}. The domain residual network (DResNet)~\cite{ma2019reducing} extends the structure of DANN~\cite{ganin2016domain} for subject-independent EEG-based vigilance estimation and emotion recognition, showcasing improved generalisation ability without target domain data. While domain adaptation methods often yield better results than domain generalisation techniques in aBCIs, an alternative approach called ASFM~\cite{chai2017fast} has been adopted, which has been integrated into an EEG-based emotion recognition setting, referred to as Plug-and-Play domain adaptation framework~\cite{zhao2021plug}.
\\
\textbf{Meta-learning:} The notion of meta-learning is to learn functional prior knowledge and involves episode-level learning~\cite{finn2017model}, has gained significant traction, particularly in the context of domain generalisation. Meta-learning for domain generalisation (MLDG)~\cite{li2018learning} presented the first meta-learning strategy to domain generalisation. Subsequently, MetaReg~\cite{balaji2018metareg} and Feature-Critic~\cite{li2019feature} were proposed to enhance the generalisation capability of the model by incorporating auxiliary losses during training. Unlike previous domain generalisation approaches that design specific models, meta-learning-based schemes focus on a model-agnostic training strategy that exposes the model to domain shifts during training.

\section{Problem Formulation}
We describe the key aspects of our problem that encompass a few components for our framework settings. The input space for EEG data is represented by $\mathcal{E}_I$, and the output space is represented by $\mathcal{E}_O$. A domain $\mathcal{D}$ is described as a joint distribution $\mathbb{P}_{\mathcal{E}_{I}\mathcal{E}_O}$ over the space $\mathcal{E}_I \times\mathcal{E}_O$. As the distribution is subject to change due to various factors, we assume that it follows a distribution $\mathcal{P}$. However, it should be noted that domains are not directly observable, and we can only observe samples $S_i$ of domains, where each $S_i$ refers to a set of $\left \{ {\mathcal{E}_{I}}_i, {\mathcal{E}_{O}}_i \right \}$. The presence of inconsistency between domains may lead to a suboptimal generalisation capability. In order to address this issue, one approach is to employ a functional mapping $\mathcal{Q}$ that transforms one domain into another while minimising the divergence between the domains. The selection of a divergence loss function $d(\cdot,\cdot)$ is typically necessary, as it considers the marginal or joint distribution. The ultimate selection of the optimal $\mathcal{Q}$ is determined by minimising the Equation~\ref{eq:eq1}.
%\vspace{-4pt}
\begin{align}\label{eq:eq1}
    \mathcal{Q}_{da}= \underset{\mathcal{Q}}{\arg\min}~d(\mathcal{Q}(S_i), \mathcal{Q}(S_j))
\end{align}
The utilisation of a functional $\mathcal{Q}_{da}$ enables the transfer of various domains to a shared feature space, thereby ensuring that the model trained on $\mathcal{Q}(S)$ does not encounter any domain shift issue. This approach is known as alignment.

\subsection{Shift-Independent Domain}
In the context of multi-source domain adaptation or domain generalisation, a common approach to incorporate domain adaptation methods involves the concurrent minimisation of the divergence between each pair of source domains, as expressed by Equation~\ref{eq:eq2}.
%\vspace{-4pt}
\begin{align}\label{eq:eq2}
    \mathcal{Q}_{mda}= \underset{\mathcal{Q}}{\arg\min}\sum_{S_i, S_j \in S_{\text{shift}}}d(\mathcal{Q}(S_i), \mathcal{Q}(S_j))
\end{align}
Under ideal circumstances, the alternative variation $\sum_{S_i, S_j \in S_{\text{shift}}}\\d(\mathcal{Q}(S_i), \mathcal{Q}(S_j))$ converges to zero, resulting in a convergence of all domains to a homogeneous state. This alternative domain, characterised by the absence of variations, is referred to as \emph{shift-independent domain}.

\begin{definition}
A shift-independent domain $\tau_{\text{shift}}$ is any $\mathcal{Q}(S_i)$ asymptotically provided $\sum_{i\neq j}d(\mathcal{Q}(S_i), \mathcal{Q}(S_j))\rightarrow 0$
\end{definition}
\begin{theorem}\label{thm1}
Given the asymptotic behaviour, the overall variation $\sum_{S_i, S_j \in S_{\text{shift}}}d(\mathcal{Q}(S_i), \mathcal{Q}(S_j))$ optimisation is identical to a loss function optimisation $\sum_{S_i \in S_{\text{shift}}}\mathcal{L}_{\mathcal{Q}(S_i)}$, where $$\mathcal{L}_{\mathcal{Q}(S_i)} := d(\mathcal{Q}(S_i), \tau_{\text{shift}})$$.
\end{theorem}
%\vspace{-18pt}
\begin{figure}[h!]
    \centering
    \includegraphics[width=5cm, height=2cm]{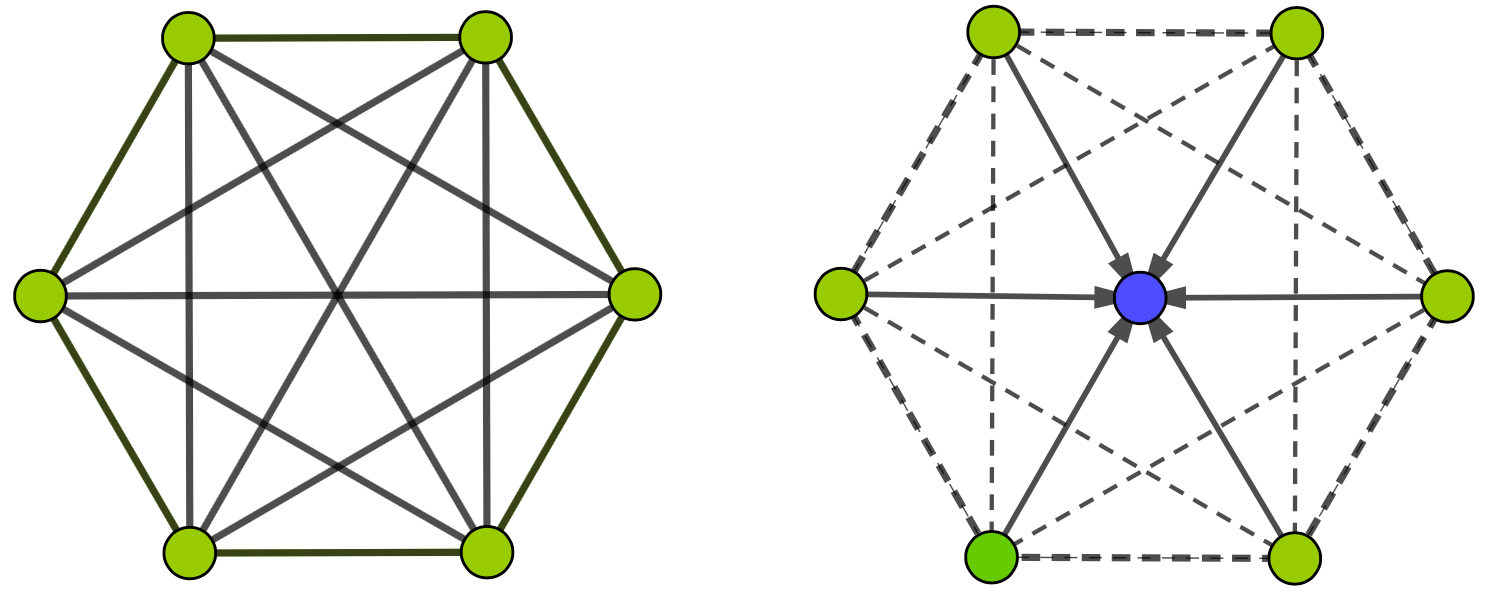}
    \caption{The schematic representation of the overall dissimilarity among pairwise domains, as well as the dissimilarity among each individual domain and the shift-independent domain.}
    \label{fig:fda}
\end{figure}
Figure~\ref{fig:fda} depicts that the simultaneous alignment of all pairwise domains can be regarded as aligning each individual domain with the shift-independent domain. Once the shift-independent domain is obtained, a network can be constructed to calculate the shift of the target domain, and
utilise it to regulate the process of meta-learning based augmented domain adaptation. In the following sections, we will elaborate on the construction of this network.

\subsection{Sum Decomposable Component}
Permutation-invariant constraints are an essential criterion for a function to capture domain discrepancy, which implies that the arrangement or order of the source domain data should not impact the output. Extensive research has been conducted on this property in prior studies~\cite{zaheer2017deep,qi2017pointnet++}. Typically, summation is commonly employed to enforce permutation invariance, leading to the concept of sum-decomposition. Definition~\ref{def2} provides a formal definition of sum-decomposable.
\begin{definition}\label{def2}
A function $f$ is said to possess sum-decomposability through $\mathbb{R}^{N}$ if there exist two functions, $\psi: \mathbb{R}\rightarrow\mathbb{R}^{N}$ and $\rho:\mathbb{R}^{N}\rightarrow\mathbb{R}$, provided $f(X)$ can be expressed as $\rho(\sum_{x\in X}\psi(X))$.
\end{definition}
\begin{theorem}\label{thm2}
A continuous map $F_m :\mathbb{R}^{M}\rightarrow\mathbb{R}$ exhibits permutation invariance if and only if it can be represented as a continuous sum-decomposition through $\mathbb{R}^{M}$.
\end{theorem}
\begin{proposition}
The established measures of domain discrepancy (or domain variation), such as Maximum Mean Discrepancy (MMD) or $\mathcal{H}$-divergence, can be derived in an equivalent manner using a function that possesses the property of sum-decomposability.
\end{proposition}
The detailed proof of Theorem~\ref{thm2} can be referred in a prior work~\cite{wagstaff2019limitations} conducted on arbitrary functions representation on sets. It is worth noting that the introduction of a summation layer or averaging layer allows for straightforward enforcement of the permutation-invariant property. Theorem~\ref{thm2} indicates that a sum-decomposable network, operating within a latent space of adequate dimensionality is capable of effectively representing any permutation-invariant function, including $d(\cdot, \tau_{\text{shift}})$ as delineated in Theorem~\ref{thm1}.

\section{The Proposed Method}
\begin{figure}[h!]
\centering%, height=9cm
  \includegraphics[width=\linewidth, height=7cm]{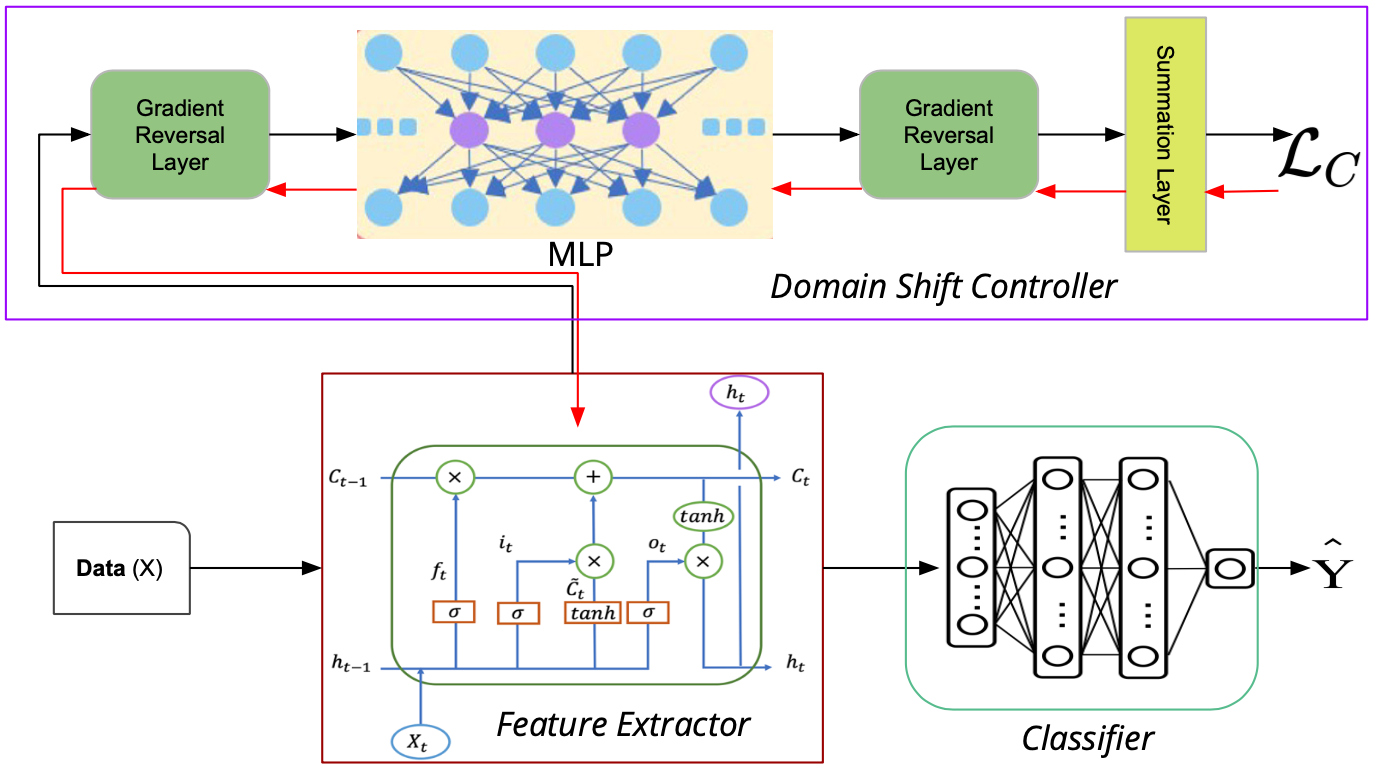}
  \caption{An overview of the proposed MeLaDA framework. The operational procedure of our framework inputs a new domain, where the long short term memory (LSTM) as a feature extractor that initially transforms the data into its corresponding feature representation. Subsequently, the controller examines the disparity between the new domain and the shift-independent domain, ultimately computing the loss function $\mathcal{L}_{C}$. Guided by the controller $D_C$, the feature extractor promptly adjusts itself through a series of adaptation steps. Furthermore, the data $X$ is forwarded to the updated feature extractor, which subsequently performs a forward propagation through the classifier.}\label{fig:arch}
\end{figure}

In accordance with the theoretical framework, our proposed approach, referred to as MeLaDA, integrates a sum-decomposable domain shift controller with the temporal multi-layer perceptron (MLP) network. The architecture, depicted in Figure~\ref{fig:arch}, consists of a feature extractor~$F(\theta)$ and a classifier~$C(\phi)$ as constituent components of the temporal MLP network. The domain shift controller $D_C (\omega)$ utilises the features extracted by $F(\theta)$ to assess the dissimilarity between the current domain and the shift-independent domain. When applying this network for the classification of target domain data, $D_C$ propagates forward to compute the domain shift and subsequently propagates backward to fine-tune the feature extractor $F$. This behaviour resembles the actions of an intelligent controller who dynamically adjusts the network based on its performance in generating shift-independent features. Under the guidance of $D_C$, the entire network exhibits the ability to generalise to unseen domains through meta-learning augmented domain adaptation. By leveraging the trained controller, the feature extractor effectively mitigates the domain shift present in the data from individual subjects. Consequently, data samples with identical emotion labels originating from diverse domains exhibit a comparable distribution within the shared space. The subsequent sections will outline the design principles for the domain shift controller and provide an overview of our training strategy.

\subsection{Model}
We describe our proposed modelling approach based on the aforementioned components. In order to satisfy the permutation-invariance requirement of the domain shift controller, a straightforward approach is to incorporate a summation layer into a neural network, similar to the method employed by Feature-Critic~\cite{li2019feature} networks. The Feature-Critic (FC) network appends a summation layer to the end of a multi-layer perceptron, disregarding the external mapping $\rho$ defined in Definition~\ref{def2}. Consequently, it may not fully embody the characteristics of a domain shift controller. Furthermore, our experimental findings indicate that introducing adversarial elements to the network can enhance its capabilities. Specifically, we incorporate two gradient reversal layers (GRL)~\cite{ganin2016domain} before and after a two-layer MLP, followed by an additional layer to further augment its performance. The rationale behind incorporating GRL into the network draws inspiration from traditional methods used to represent domain discrepancy, such as MMD or adversarial-based approaches~\cite{ganin2015unsupervised,shen2018wasserstein}. These methods share a common principle, which entails utilising the ``largest" difference between two domains to depict their divergence. To imbue our network with the same capacity for simulating domain shift as these traditional methods, we introduce a novel divergence measure, akin to MMD and $\mathcal{H}$-divergence, that we term maximum mean norm discrepancy ($M_{\text{MND}}$).
%\vspace{-8pt}
%\vspace{-4pt}
\begin{definition}
Given the maximum mean norm discrepancy~$M_{\text{MND}}$ among two domains $S_i$ and $S_j$ is characterised as 
\vspace{-6pt}
\begin{align}\label{eq3}
M_{\text{MND}}(S_i, S_j) := \max_{g\in\mathcal{Q}}\left \|\mathbb{E}_{x\in S_{i}}g(x)-\mathbb{E}_{x\in S_{j}}g(x)  \right \|_2
\end{align}
, where a function $g$ maps $x$ into a vector space.
\end{definition}
%\vspace{-6pt}
Based on Theorem~\ref{thm1}, the selection of an implicit domain as the shift-independent domain is a viable approach to circumvent the need for direct domain comparison. The implicit domain choice aims to minimise the overall divergence, facilitating efficient optimisation. Consequently, our proposed objective function encompasses both minimisation and maximisation components, as indicated by Equation~\ref{eq4},
\vspace{-8pt}
\begin{align}\label{eq4}
    \mathcal{L}_{c} =\sum_{S_i \in S_{\text{shift}}}\max_{g\in\mathcal{Q}}\min_{\tau\in V_{\text{lspace}}}\left \|\mathbb{E}_{x\in S_{i}}g(x)-\tau  \right \|_2
\end{align}
where the loss function of the controller's output is represented by $\mathcal{L}_{c}$. Within the domain shift controller framework, the function $g$ corresponds to the inner mapping $\psi$ in Equation~\ref{eq:eq2} and is represented by the initial layers of the network. The subsequent summation layer calculates the mean value of $g(x)$, while the final layer of the domain shift controller computes the norm of the difference. To address the maximisation and minimisation objectives, we employ the method proposed by GAN~\cite{goodfellow2020generative}, which involves the inclusion of gradient reversal layers in the controller network. During forward propagation, the GRL operates as an identity map, but during backward propagation, it reverses the direction of gradients. In contrast to the domain adaptation regulariser utilised in domain-adversarial networks~\cite{ganin2015unsupervised}, our domain shift controller incorporates an additional minimisation task. Consequently, two GRL layers are employed, the left GRL facilitates the controller into identifying the most significant domain difference, while the right GRL adjusts the feature extractor to generate consistent features across varied domains. It is worth noting that the two gradient reversal layers structure may exhibit instability during experiments. To mitigate this, we employ a strategy where we freeze a portion of the network between the two GRL layers once a predefined iteration threshold is reached, ensuring stability in the training process.
%\vspace{-2pt}
\subsection{Meta-learning based Training Strategy}
In this section, we present a meta-learning based strategy for the parameter learning process. In order to ensure the domain shift controller's ability for generalisation, we formulate the algorithmic approach for training settings. The overall algorithm can be divided into two distinct components, the training of the domain shift controller and the training of the network. These components are delineated in Algorithm~\ref{alg:controller} and Algorithm~\ref{alg:enet}, respectively.
\\
\textbf{Training Procedure for the Domain Shift Controller:} In order to augment the model's capability for generalisation, we adopt a two-fold approach that involves optimising the output $\mathcal{L}_c$ of the domain shift controller to mitigate domain shift, while simultaneously incorporating meta-learning to facilitate the generalisation process. Initially, the available domains are randomly divided into meta-train domains denoted as $S_{\text{train}}$ and meta-test domains, alternatively referred to as meta-validation domains, denoted as $S_{\text{valid}}$. The controller is leveraged to optimise the feature extractor $F(\theta)$ specifically on the meta-train domains. Subsequently, we assess the efficacy of the optimised feature extractor $F(\theta^{'})$ on the meta-test domains, evaluating its performance and ability to generalise beyond the trained domains. After updating the parameter $\theta$ to $\theta^{'}$, the classification loss function, represented by $\ell$, undergoes a transformation from $\ell(x^{i},y^{i};\theta)$ to $\ell(x^{i},y^{i};\theta^{'})$. Inspired by the approach developed in Feature-critic networks~\cite{li2019feature}, we establish the meta loss function as depicted in Equation~\ref{eq5}.
%\vspace{-6pt}
\begin{align}\label{eq5}
    \mathcal{L}_{\text{meta}}=\sum_{(x_i,y_i)\in S_{\text{valid}}}\tanh\left (\ell(x^{i},y^{i};\theta^{'})-(\ell(x^{i},y^{i};\theta) \right )
\end{align}
The overall loss for updating the parameter $\omega$ is expressed as depicted in Equation~\ref{eq6}.
\vspace{-6pt}
\begin{align}\label{eq6}
\mathcal{L}_C(\theta,\phi,\omega;S_{\text{train}})+\lambda\mathcal{L}_{\text{meta}}(\theta^{'},\phi,\omega;S_{\text{valid}})
\end{align}
The hyperparameter $\lambda$, as well as the parameters $\theta$, $\phi$, and $\omega$ corresponding to $F$, $C$, and $D_C$ respectively, are involved in Equation~\ref{eq6}. Through the optimisation of Equation~\ref{eq6}, the parameter $\omega$ of $D_C(\omega)$ is ultimately updated accordingly.
\\
\textbf{Training Procedure for the MeLaDA framework:} In contrast to the training approach employed by MetaReg~\cite{balaji2018metareg} or Feature-Critic networks~\cite{li2019feature}, which involves training the auxiliary network before training the task network, our proposed method MeLaDA adopts an alternative training scheme for the domain shift controller and temporal MLP network. This is necessary because the controller network~$D_C$ needs to remain functional during the test phase, requiring continuous updates even while other parts of the network are being trained. Additionally, to fully adhere to the principles of meta-learning, we leverage the model-agnostic meta-learning (MAML)~\cite{finn2017model} framework to train the network, rather than directly optimising it. We treat the domain shift controller and classification as two distinct tasks and utilise episodic training~\cite{li2019episodic} to update their respective parameters. The dataset is divided into two subsets, $S_{\text{train}}$ (meta-train domains) and $S_{\text{valid}}$ (meta-validation domains). The domain shift controller, $D_C$, utilises data from $S_{\text{train}}$ to compute the domain shift loss, $\mathcal{L}_C(\theta,\omega;S_{\text{train}})$, as well as the classification loss, $\mathcal{L}^{\text{train}}_{\text{classif}}$. These losses are then used to update the parameters of the feature extractor, $F(\theta)$. Subsequently, with the updated parameters, $F(\theta^{'})$, processes data from $S_{\text{valid}}$, and the temporal MLP network computes the corresponding classification loss, $\mathcal{L}^{\text{valid}}_{\text{classif}}$. The overall loss function for optimising $F(\theta)$ and $C(\phi)$ is defined as follows
%\vspace{-6pt}
\begin{align}\label{eq7}
\lambda\mathcal{L}_C(\theta,\omega;S_{\text{train}})+\mathcal{L}^{\text{train}}_{\text{classif}}(\theta,\phi)+\mathcal{L}^{\text{valid}}_{\text{classif}}(\theta^{'},\phi)
\end{align}

%\vspace{-10pt}
\begin{algorithm}
\caption{Training the Domain Shift Controller}\label{alg:controller}
\begin{algorithmic}[1]
\Require Given a domain \(\mathcal{D}\) and \(\theta\), \(\phi\), \(\omega\) are parameters.
\Ensure \(\omega\)
\State  $\mathcal{D}:(S_{\text{train}}, S_{\text{valid}})\leftarrow\mathcal{D}$ \Comment{Random partitioning}
\State $\mathcal{L}_C(\theta,\omega;S_{\text{train}})\leftarrow D_{C}(F(X))$ \Comment{Meta-training stage}
\State $\theta\leftarrow\theta^{'}-\alpha\triangledown_{\theta}\mathcal{L}_C(\theta,\omega;S_{\text{train}})$ \Comment{Meta-training stage}
\State $\mathcal{L}_{\text{meta}}(\theta,\theta^{'},\phi;S_{\text{valid}})\leftarrow$ \Comment{Meta-validation stage}
\WRP $\sum_{(x_i,y_i)\in S_{\text{valid}}}\tanh\left (\ell(x^{i},y^{i};\theta^{'})-(\ell(x^{i},y^{i};\theta) \right)$
\State Update $\omega$ utilising $\mathcal{L}_C +\lambda\mathcal{L}_{\text{meta}}$ \Comment{Optimisation}
\end{algorithmic}
\end{algorithm}
\vspace{-10pt}
\begin{algorithm}
\caption{Training the MeLaDA framework}\label{alg:enet}
\begin{algorithmic}[1]
\Require Given a domain \(\mathcal{D}\) and \(\theta\), \(\phi\), \(\omega\) are parameters.
\Ensure \(\theta\), \(\phi\), \(\omega\)
\State  $\mathcal{D}:(S_{\text{train}}, S_{\text{valid}})\leftarrow\mathcal{D}$ \Comment{Random partitioning}
\State $\mathcal{L}_C(\theta,\omega;S_{\text{train}})\leftarrow D_{C}(F(X))$ \Comment{Meta-training stage}
\State $\mathcal{L}^{\text{train}}_{\text{classif}}(\theta,\phi)\leftarrow \ell(C(F(X_{\text{train}})),Y_{\text{train}})  $ \Comment{Meta-training stage}
\State $\theta\leftarrow\theta^{'}-\alpha\triangledown_{\theta}\mathcal{L}_C(\theta,\omega;S_{\text{train}})$ \Comment{Meta-training stage}
\State $\mathcal{L}^{\text{valid}}_{\text{classif}}(\theta^{'},\phi)\leftarrow \ell(C(F(X_{\text{valid}})),Y_{\text{valid}})$\WRP\Comment{Meta-validation stage}
\State Update $\theta, \phi, \omega$ utilising $\lambda\mathcal{L}_C +\mathcal{L}^{\text{valid}}_{\text{classif}}+\mathcal{L}^{\text{train}}_{\text{classif}}$\WRP\Comment{Optimisation}
\end{algorithmic}
\end{algorithm}
The introduction of MeLaDA adopts an alignment-based domain adaptation perspective. However, an alternative explanation of this method can be provided through the lens of meta-learning. Recent domain generalisation approaches~\cite{li2018learning} suggest that meta-learning involves linking various tasks by aligning their gradients in a shared direction. In our scenario, the domain shift controller task is deliberately designed to be coupled with the ``domain generalisation" process. Consequently, when the model encounters a new domain, the optimisation objective of adapting to this domain aligns with the optimisation objective propelled by the controller.

\section{Experiments}
\subsection{Dataset and Feature Extraction}
In our evaluation of the MeLaDA framework, we employ the SEED dataset~\cite{zheng2015investigating} which was created for emotion recognition and aBCIs using EEG signals. This dataset encompasses EEG signals collected from 15 subjects who were enlisted to watch carefully curated 4 minutes of film clips. These clips were specifically chosen to elicit one of three distinct emotions which are happiness, neutrality, and sadness. Each subject have been subjected to an experiment 3 times in intervals of one week. During the selection process of film clips, stringent criteria were applied to ensure that each clip was well-edited, enabling the creation of coherent emotion elicitation while maximising emotional significance. The EEG signals were recorded using the ESI NeuroScan system, employing a 62-electrode headset. The sampling rate for the signals was set at 1000 Hz. 
By utilising the SEED dataset, we were able to assess the performance and effectiveness of our MeLaDA approach in the context of emotion recognition. The dataset's comprehensive nature and carefully designed stimuli provide a valuable resource for training, testing, and validating algorithms and models aimed at understanding and interpreting emotions from EEG signals. The feature extraction process follows the similar strategies by deep belief networks for EEG-driven emotion recognition~\cite{zheng2015investigating}. Given that the SEED dataset has already undergone preprocessing, we are able to directly extract the features. Specifically, we employ the differential entropy feature~\cite{duan2013differential}, which has been previously shown to be effective for EEG-based emotion recognition in several studies~\cite{zheng2015investigating,zheng2017identifying}. Existing work~\cite{shi2013differential} have demonstrated that the differential entropy feature corresponds to the logarithmic spectral energy of a fixed-length EEG sequence within a specific frequency band. To obtain the spectral energy, we apply the short-time Fourier transform using a non-overlapping Hanning window of 1 second to the EEG signal, considering five frequency bands, which are $\delta$ ranging from 1 Hz to 3 Hz, $\theta$ from 4 Hz to 7 Hz, $\alpha$ from 8 Hz to 13 Hz, $\beta$ from 14 Hz to 30 Hz, and $\gamma$ from 31 Hz to 50 Hz. Subsequently, we compute the differential entropy feature. Considering the inherent dynamism observed in EEG-based emotion recognition tasks, we integrate the linear dynamic system methodology to effectively filter the differential entropy feature. Each sample has a dimension of 310 (62 channels $\times$ 5 frequency bands). Since the EEG data consist of time series, we resample the feature with a time-step of 15 and a 1-second overlap, resulting in 3184 samples per subject.
\vspace{-12pt}
\subsection{Parameter and Implementation Settings}
In line with the Plug-and-play (PnP) approach~\cite{zhao2021plug}, we have adopted the leave-one-subject-out (LOSO) strategy to assess the generalisation capability of the MeLaDA framework. For each iteration, one subject is selected as the target, while the remaining 14 subjects are used to train our model. During the test phase, the prediction results obtained after 10 steps of self-adaptation are utilised. The feature extractor component of MeLaDA consists of a two-layer LSTM network with an output dimension of 256 and a time step of 15. The classifier is implemented as a two-layer MLP with a hidden size of 100. Both the temporal MLP network and the domain shift controller undergo optimisation using the Adam optimizer with a learning rate of 0.0002 and a weight decay of 0.0001. The parameter $\lambda$ is assigned a value of 0.1. Initially, the temporal MLP network is pre-trained until it achieves an accuracy of over 85\% on the training set. Subsequently, MeLaDA is employed to jointly train the domain shift controller and the temporal MLP network. The threshold for freezing a portion of the controller within the gradient reversal layers is set to 40, and the maximum number of iterations is set to 200.
\vspace{-8pt}
\section{Results and Discussion}
To assess the efficacy of our proposed MeLaDA framework, we employ the leave-one-subject-out cross-validation evaluation scheme and conduct a comparative analysis between MeLaDA and various domain adaptation and domain generalisation approaches using the SEED dataset. The evaluation results, comprising the mean accuracy (MA) and standard deviation (SD), are presented in Table 1. In contrast to the baseline approach, which involves aggregating data from all source domains and training a single model using the support vector machine (SVM), all the evaluated methods exhibit a significant improvement in accuracy of at least 13\%. Notably, MeLaDA surpasses all domain generalisation methods in terms of performance. When compared to the domain adaptation methods, MeLaDA still achieves commendable results. Although WGAN-DA~\cite{wganda} and Plug-and-Play method~\cite{zhao2021plug} exhibit marginally higher accuracy than MeLaDA. It should be noted that WGAN-DA requires all source domains and the Plug-and-Play method necessitates the utilisation of a subset of domains for adaptation, thereby limiting the fast generalisation capability of PnP method.
\begin{table}[btp]
   \centering 
   \begin{tabular}{|c|c|c|c|c|c|}
    \hline 
       \multirow{1}{*}{\textbf{Models}} &
       \multicolumn{2}{c|}{\textbf{DA}} & 
       \multirow{1}{*}{\textbf{Models}} &
       \multicolumn{2}{c|}{\textbf{DG}} \\
      & MA & SD & & MA & SD \\
      \hline
      SVM~\cite{zheng2016personalizing} & 0.567 & 0.16 & ~-~ & ~-~ & ~-~ \\
      \hline
      TCA~\cite{zheng2016personalizing} & 0.640 & 0.15 & MLDG~\cite{li2018learning} & 0.795 & 0.12 \\
      \hline
      TPT~\cite{zheng2016personalizing} & 0.752 & 0.13 & FC~\cite{li2019feature} & 0.806 & 0.12 \\
      \hline
      DAN~\cite{li2018cross} & 0.838 & 0.08 & DICA~\cite{ma2019reducing} & 0.7 & 0.08 \\
      \hline
      DANN~\cite{li2018cross} & 0.792 & 0.13 & DResNet~\cite{ma2019reducing} & 0.853 & 0.08\\
      \hline
      WGAN-DA~\cite{wganda} & 0.871 & 0.07 & PnP~\cite{zhao2021plug} & 0.854 & 0.07\\
      \hline
      \cellcolor{green}\textbf{MeLaDA}~(Ours) & \cellcolor{green}0.864 & \cellcolor{green}0.09 & \cellcolor{green}MeLaDA & \cellcolor{green}0.864 & \cellcolor{green}0.09\\
      \hline
   \end{tabular}
   \caption{\label{tab:melada}The mean accuracy and standard deviation for both domain adaptation (DA) and domain generalisation (DG) are reported with comparative baseline methods on the SEED dataset.}
\end{table}
It is important to note that our proposed method, being an implementation of a meta-learning strategy, achieves a superior accuracy compared to MLDG~\cite{li2018learning} and Feature-Critic~\cite{li2019feature} by approximately 7\% and 6\% respectively. These results indicate that MLDG, which directly employs episodic training to generalise the model is insufficient in effectively addressing the subject variability inherent in EEG-based emotion recognition. Similarly, the Feature-Critic network, despite utilising a sum-decomposable MLP to simulate domain shift during the training phase, does not lead to a significant improvement in the results. This suggests that the application of a domain shift controller during the testing phase proves to be beneficial. As an augmented domain adaptation method, MeLaDA demonstrates the capability to predict any target set with only a few steps of self-adaptation, thereby leveraging the advantages of both domain adaptation and domain generalisation.
\begin{figure}[hbt!]
    \centering
    \includegraphics[width=\linewidth,height=4cm]{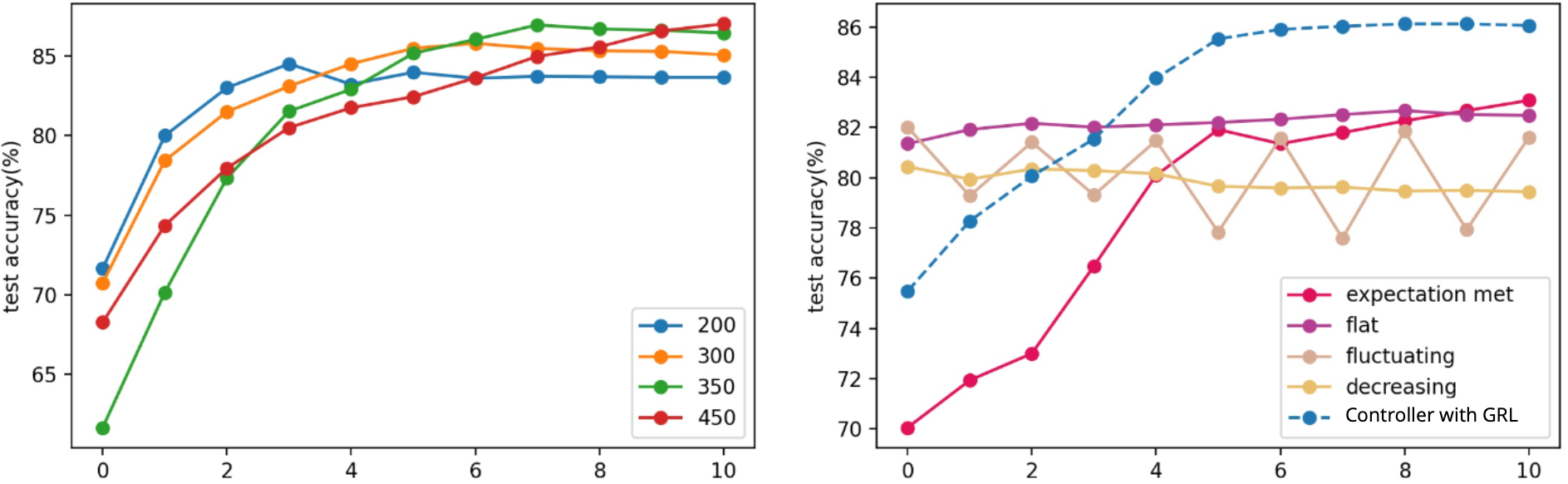}
    \caption{Results of self-adaptation with the left subplot delineating the test accuracy of MeLaDA given the X-axis is the number of iterations. The right subplot represents the test accuracy of MeLaDA without gradient reversal layers.}
    \label{fig:ablation}
\end{figure}
Also, we examine the domain shift controller performance by measuring the accuracy of self-adaptation during the test phase. Figure~\ref{fig:ablation} demonstrates the self-adaptation capabilities of MeLaDA when confronted with a new domain. Remarkably, the model achieves favourable performance within a limited number of self-adaptation steps. The left subplot illustrates the adaptation performance on the target domain during the training phase, where each self-adaptation process relies solely on the input data for prediction without requiring any additional information. The right subplot compares the performance of the controller with and without the GRL. This indicates that the inclusion of the adversarial strategy enhances the stability and efficiency of the domain shift controller, as evidenced by reduced fluctuations and improved overall performance, whereas the absence of GRL may result in fluctuations or performance deterioration.

\section{Conclusion}
In this study, we present an augmented domain adaptation approach, MeLaDA for dealing with a subject-agnostic model for EEG-based emotion recognition without the need for source domain data in the test phase. Our proposed approach adopted a sum-decomposable domain shift controller to facilitate augmented domain adaptation. By integrating adversarial learning and meta-learning techniques, MeLaDA demonstrates the ability to generalise to new domains with minimal self-adaptive iterations. Experimental results conducted on the SEED dataset showcase the superiority of MeLaDA over traditional domain generalisation methods in terms of performance. This highlights the suitability of MeLaDA for constructing subject-agnostic affective models, surpassing conventional domain adaptation, domain generalisation, and ASFM methods.

\begin{comment}
\begin{acknowledgments}
  Thanks to the developers of ACM consolidated LaTeX styles
  \url{https://github.com/borisveytsman/acmart} and to the developers
  of Elsevier updated \LaTeX{} templates
  \url{https://www.ctan.org/tex-archive/macros/latex/contrib/els-cas-templates}.  
\end{acknowledgments}
\end{comment}
%%
%% Define the bibliography file to be used
\bibliography{sample-ceur}

%\appendix


%The sources for the ceur-art style are available via
\begin{comment}
\begin{itemize}
\item \href{https://github.com/yamadharma/ceurart}{GitHub},
% \item \href{https://www.overleaf.com/project/5e76702c4acae70001d3bc87}{Overleaf},
\item
  \href{https://www.overleaf.com/latex/templates/template-for-submissions-to-ceur-workshop-proceedings-ceur-ws-dot-org/pkfscdkgkhcq}{Overleaf
    template}.
\end{itemize}
\end{comment}
\end{document}